# COMPARATIVE STUDY OF DATA MINING QUERY LANGUAGES

Mohamed Anis Bach Tobji
*LARODEC Laboratory – Institut Supérieur de Gestion de Tunis*
*41 rue de la liberté Bouchoucha, 2000, Tunis - Tunisia*

**ABSTRACT**

Since formulation of Inductive Database (IDB) problem, several Data Mining (DM) languages have been proposed, confirming that KDD process could be supported via inductive queries (IQ) answering. This paper reviews the existing DM languages. We are presenting important primitives of the DM language and classifying our languages according to primitives' satisfaction. In addition, we presented languages' syntaxes and tried to apply each one to a database sample to test a set of KDD operations. This study allows us to highlight languages capabilities and limits, which is very useful for future work and perspectives.

## 1. INTRODUCTION

IDB is a new generation of databases introduced in (Imielinski and Mannila, 1996) as a framework of KDD (Fayyad et al, 1996). An inductive database contains data and patterns that are extracted from. Databases are generally supported by SQL language, however, IDBs are supported by a DM Query Language, which allows KDD operations (mainly data selection, data preprocessing, patterns mining and pattern post-processing).

The development of theoretical framework is interesting and has been the subject of many researches (Boulicaut et al, 1999), (De Raedt, 2003), (Dan Lee and De Raedt, 2003), (De Raedt et al, 2004). However, there is no clear definition or formalization, such as an algebra language that could be a base for a standard DM query language. In fact, the KDD community would reproduce the success of SQL based on Codd's algebra (Codd, 1970).

In this paper we study existing DM languages to try to find out advantages and limits. The paper is organized as the following: In section 2 we present essential DM query language primitives. In section 3 we compare six existing DM query languages with a taxonomy based on primitives' satisfaction. In section 4, we show the languages in action, i.e., we give a small database and we perform some data mining operations using languages' queries. Finally in section 5, we discuss the study, and we give some perspectives related to the existent languages weaknesses.

## 2. INDUCTIVE QUERY LANGUAGE PRIMITIVES

Data mining query language primitives' definition is a basic problem. Once primitives are defined, conceiving a good DM query language will be easier. In this section we give the primitives as defined in (Han and Kamber,2000), (Botta et al, 2004), and languages papers (Imielinski and Virmani, 1999), (Meo et al, 2002), (Han et al,1996), (Morzy and Zakrzewic, 1997), (Netz et al,2000) and (Elfeky et al, 2000).

A data mining query language must offer:
- *Data selection*: it's naturally satisfied if the language nests SQL. The language must provide data selection query.
- *Pre-processing task*: providing pre-processing operations (sampling, discretization, data cleaning etc. )
- *Specifying the data mining task*: mining several patterns kinds (decision trees, sequential and association rules etc.).

- *Specification of background knowledge*: background knowledge is information about the application field. This primitive offers to the Data Miner the opportunity to specify his domain knowledge which affects positively the mined knowledge quality. Concept hierarchy is the most used background knowledge (Han and Kamber,2000).
- *Specification of constraints mining*: specification of constraints set that the patterns must satisfy.
- *Closure property*: the result of data mining query could be re-queried such as for SQL.
- *Post-processing task*: the user should be able to query extracted patterns, cross over patterns and data etc.

## 3. THE COMPARATIVE TABLE

In this section, we study six DM query languages. We present these languages according to a set of properties corresponding to the primitives defined in the previous section. Thus, we classify the languages in a table such that rows correspond to properties and columns to languages. Eeach cell (crossing a property $Pi$ and a language $Lj$) is the satisfaction degree of the property $Pi$ by the language $Lj$ (see table 1).

Table 1 contains two parts. In the first one, each language is described generally (language authors, design, year etc). In the second part, we present the functionalities provided by each language as explained on the top.

## 4. DATA MINING QUERY LANGUAGES IN ACTION

In this section, we explore DM query languages capabilities and we present the syntax of each language. In addition, we set a database example about supermarket sales (see table 2) and tried to write some queries to perform KDD operations that turn around the DM step, mainly in order to extract association rules since their mining is provided by all the languages.

### 4.1. MSQL

MSQL has four main queries:
- `Create Encoding`: this query allows discretization of continuous attributes. The command creates ranges of values, and assigns discrete integers to those ranges. Discretization is done "on the fly" and discrete values are not stored anywhere.
- `GetRules`: this query allows the generation of association rules, which are stored in a suitable table.
- `SelectRules`: this query allows the selection of association rules. This is very important to post-process the mined patterns.
- `Satisfies and Violates`: these operators allow cross-over between data and association rules: that is selecting data generating a set of association rules (satisfies), or data that contradict (violates) them.

**Illustration**

In this subsection (repeated for each language section), we tried to mine association rules with bodies about items and head about income customer from sales database example (see table 2). We choose to mine only transactions paid by credit card, thus extracted associations are only about this kind of transactions.

MSQL needs a special data format to mine association rules; MSQL works in a way that each descriptor is an attribute. The domain of an attribute should be a set of discrete values. Thus, continuous attributes have to be encoded, i.e., discretized. Table3 represents the first row of our input relations (items bought, payment mode and customer income).

This transformation is manually done by the user since there is no MSQL transformation query.

- **Pre-processing**

First of all, we have to select data we want to mine. In our example, we want to mine only transactions paid by credit cards for associations between *items* and *income*. Thus, we create a view named *transaction_view* using a classical SQL query that extracts data from the relation transactions:

```
CREATE VIEW TRANSACTION_VIEW AS
SELECT ID_TRANSAC,INCOME,PAYMENT_MODE, SUM(CASE ITEM WHEN 'A' THEN 1 ELSE 0 END)) AS S /*repeat the
SUM(CASE… expression for each item in the table transactions*/
FROM TRANSACTIONS T, CUSTOMER C
WHERE T.ID_CUSTOMER=C.ID_CUSTOMER AND PAYMENT_MODE='Credit_card'
GROUP BY ID_CUSTOMER,ID_TRANSACTION,INCOME,PAYMENT_MODE;
```

In the second step, we must discretize the attribute income which is continuous. This query allows the discretization of the attribute income. Discrete values are not materialized:

```
CREATE ENCODING DSCRT_INCOME ON TRANSACTION_VIEW.INCOME AS
```

```
BEGIN
(MIN,499,1), (500,599,2), (600,699,3), (700,799,4), (800,MAX,5), 0
END;
```

Table 1. Languages comparative table

| Properties | | Languages | MSQL | MINE RULE | DMQL | MineSQL | DMX | ODMQL |
|---|---|---|---|---|---|---|---|---|
| **General Description** | | | | | | | | |
| Year | | | 1996 | 1996 | 1996 | 1997 | 2000 | 2000 |
| Authors | | | Imielinski and Virmani | Meo, Psaila and Ceri | Han, Fu, Wang, Koperski and Zaiane | Morzy and Zakrzewicz | Netz, Chaudhuri, Bernhardt and Fayyad | Elfeky, Saad and Fouad |
| **Primitives Satisfaction** | | | | | | | | |
| Data Selection | | | No | Yes | Yes | Yes | Yes | Yes |
| Specification of the pattern type | Association rules | One item head | Yes | Yes | Yes | Yes | Yes | Yes |
| | | More than one | No | Yes | Yes | Yes | Yes | Yes |
| | Characteristic rules | | No | No | Yes | Yes | No | Yes |
| | Discriminant rules | | No | No | Yes | No | No | Yes |
| | Classification rules | | No | No | Yes | No | Yes | Yes |
| | Sequential patterns | | No | Yes | No | No | Yes | No |
| Background knowledge specification | | | No | Some (Item hierarchy) | Some (concept hierarchies) | Some (concept hierarchies) | Some (concept hierarchies) | Some (concept hierarchies) |
| Constraints definition | Choice of significance measures | | No | No | Yes | No | Yes | No |
| | Syntactic constraints | | Yes | Yes | Yes | Yes | Yes | No |
| | Optimization constraints | | Some (maximal rules) | No | No | Some (thanks to some functions) | Some | No |
| Closure property | | | Yes | Yes | Yes | Yes | Yes | Not defined |
| Pre-processing | | | Some (Discret. of continuous attributes) | No | No | No | Yes | No |
| Crossing over patterns and data | | | Yes | No | No | Yes | Yes | No |
| Post-processing | | | Yes | No | Some (Patterns visualization) | Yes | Some | No |

Table 2. Sales database (on the left table *transactions*, on the middle table *customer,* on the right description of attributes).

| id_customer | id_transac | item | Payment_mode |
|---|---|---|---|
| c1 | t1 | AB | credit_card |
| c2 | t2 | CD | credit_card |
| c3 | t3 | CD | cash |
| c4 | t4 | E | credit_card |
| c5 | t5 | CE | credit_card |
| c6 | t6 | BD | cash |
| c7 | t7 | ABD | credit_card |
| c8 | t8 | ABDE | credit_card |
| c9 | t9 | B | credit_card |
| c3 | t10 | A | credit_card |
| c2 | t11 | ADE | cash |

| id_customer | income |
|---|---|
| c1 | 534 |
| c2 | 668 |
| c3 | 716 |
| c4 | 659 |
| c5 | 737 |
| c6 | 563 |
| c7 | 591 |
| c8 | 525 |
| c9 | 573 |
| *id_customer* | the purchaser identifier |
| *id_transac* | the transaction identifier |
| *item* | the item bought in one transaction |
| *payment_mode* | the payment way |

Table 3 first row of our suitable relation *transactions*

| Id_transac | A | B | C | D | E | income | Payement_mode |
|---|---|---|---|---|---|---|---|
| t1 | 1 | 1 | 0 | 0 | 0 | 534 | Credit_card |

- **Association Rule Mining**

In this step, we want to mine association rules with the following form: *items→income*. Association rules support and confidence are fixed respectively to 2 and 0.5. To achieve this task, we write this MSQL query:

```
GETRULES(TRANSACTION_VIEW) INTO TRANSACTION_RB
WHERE BODY HAS {(A=1) OR (B=1) OR (C=1) OR (E=1)} AND
      CONSEQUENT IS{(INCOME=*)} AND SUPPORT>2 AND CONFIDENCE>=0.5
```

```
USING DSCRT_INCOME FOR INCOME;
```
The resulting association rules are stored in the table *transaction_rb* (see table 4).

- **Post-processing**

In this step, we try to cross-over data and patterns mined in the previous step, assuming we want to extract data that violate all association rules and whose bodies have the item A; this is the suitable query:
```
INSERT INTO CROSS_OVER AS
SELECT * FROM TRANSACTION_VIEW
WHERE VIOLATES ALL (SELECTRULES(TRANSACTION_RB) WHERE BODY HAS {(A=1)});
```
The table *CROSS_OVER* is composed of transactions t2, t4, t5, t9 and t10; these tuples violates every rule having A in the body.

Table 1 : Table transaction_rb

| Body | Consequent | Support | Confidence |
|---|---|---|---|
| A=1 | income=[500,599] | 3 | 75% |
| B=1 | income=[500,599] | 4 | 100% |
| D=1 | income=[500,599] | 3 | 66% |
| A=1^B=1 | income=[500,599] | 3 | 100% |

In the same step, we also try to select rules having the maximal body for the same head, i.e., selection of rules with bodies included in no other rule having the same head.
```
SELECTRULES(TRANSACTION_RB) AS R1
WHERE NOT EXISTS(SELECTRULES(TRANSACTION_RB) AS R2
                 WHERE R2.BODY HAS R1.BODY
                 AND NOT (R2.BODY IS R1.BODY)
                 AND R2.CONSEQUENT IS R1.CONSEQUENT);
```
We have two rules satisfying our conditions, i.e., having the maximal body (D=1)→ revenu=[500,599] and (A=1)^(B=1)→ revenu=[500,599].

## 4.2. DMQL

DMQL defines an association rule as a relationship between two predicates sets. In a predicate P(X,v), *P* is an attribute, *X* is a variable and *v* is a value belonging to the attribute domain. Thus, the association rule AB→CD is equivalent to item(X,A)^item(X,B)→item(X,C)^item(X,D). The main DMQL query allows mining patterns. Here, we present its syntax:
```
USE DATABASE (DATABASE_NAME)
{USE HIERARCHY (HIERARCHY_NAME) FOR (ATTRIBUTE)}
FIND ASSOCIATIONS [AS (RULE_NAME)]
FROM (RELATION(S))
[WHERE (CONDITION)]
[ORDER BY (ORDER_LIST)]
{WITH [(KINDS_OF)] THRESHOLD=(THRESHOLD_VALUE)}
```
This query extracts association rules in the table *rule_name*, under the thresholds of the WITH clause, from the database *database_name* and especially relations of the FROM clause filtered by the WHERE clause.

**Illustration**

In this section, we try to re-do the same operations performed for MSQL.

- **Pre-processing**

With DMQL, it's possible to discretize continuous attributes using the concept hierarchy definition query:
```
DEFINE HIERARCHY INCOME_HIERARCHY FOR INCOME ON TRANSACTION AS
LEVEL1 :{MIN…499}$<$LEVEL0 :ALL
LEVEL1 :{500…599}$<$LEVEL0 :ALL
LEVEL1 :{600…699}$<$LEVEL0 :ALL
LEVEL1 :{700…799}$<$LEVEL0 :ALL
LEVEL1 :{800…MAX}$<$LEVEL0 :ALL
```
- **Association Rule Mining**

DMQL allows syntactic constraints specification using *metapaterns*. The metapaterns are used to focus the discovery towards patterns that match given templates. Here, we want to mine rules that include in the body different items and in the head the income, with a support fixed to 25% and a confidence fixed to 50%:
```
USE DATABASE SALES
USE HIERARCHY INCOME_HIERARCHY FOR INCOME
FIND ASSOCIATIONS AS TRANSACTION_RB
MATCHING WITH ITEM+(X,{I}) → INCOME(X,A)
FROM TRANSACTIONS T, CUSTOMER C
WHERE T.ID_CUSTOMER=C.ID_CUSTOMER AND PAYMENT_MODE='credit_card'
GROUP BY ID_TRANSAC
```

```
WITH SUPPORT THRESHOLD=25%
WITH CONFIDENCE THRESHOLD=50%;
```
In metapaterns, symbol '+' means that we search rules whose bodies are composed of one or more items.

In table 5, we present association rules extracted by the previous query. We distinguish three tables: the right one, the middle one and the left one are used respectively to represent rules heads, rules bodies and relationships between the two rules components tables.

Table 2 : DMQL association rules representation

| id_r | id_b | id_h | support | confidance |
|------|------|------|---------|------------|
| 1    | 1    | 5    | 37.5%   | 75%        |
| 2    | 2    | 5    | 50%     | 100%       |
| 3    | 3    | 5    | 37.5%   | 66%        |
| 4    | 4    | 5    | 37.5%   | 100%       |

| id_b | item |
|------|------|
| 1    | A    |
| 2    | B    |
| 3    | D    |
| 4    | A    |
| 4    | B    |

| id_h | income    |
|------|-----------|
| 5    | [500,599] |

- **Post-processing**

DMQL doesn't provide a cross-over query. Post-processing operations are feasible via SQL queries on rule tables. Theses queries are very complex, it is not explicit for novice SQL users.

We tried to write queries (1) that extract data that violates all rules having the item 'A' in their bodies and (2) that select all maximal rules. We tried to simplify queries written in [BBM+02] but the result remains complex to novice SQL users. To make queries easier, we create the view RULES which is the natural joint of our three rules tables, and the view SALES which the natural joint of tables TRANSACTIONS and CUSTOMER.

```
SELECT DISTINCT id_transac FROM sales /*We select all data and we*/
MINUS /*subtract from them all data that satisfy at least one rule*/
SELECT DISTINCT id_transac FROM sales S /*with A in the body.*/
WHERE EXISTS(SELECT id_r FROM rules R /*we search data such*/
WHERE item='A' /*it exists at least one rule which*/
/*is included*/ AND NOT EXISTS(SELECT ITEM,INCOME FROM rules Q
/*in these data. A set X is*/     WHERE R.id_r=Q.id_r
/*included in another one Y*/     MINUS
/*if X minus Y is the empty set.*/SELECT item,income FROM sales T
  where T.id_transac=S.id_transac));
```

In order to obtain data that violates all rules having 'A' in the body, we subtract data that satisfy at least one rule having 'A' in the body from all the data. Now, we select maximal rules bodies.

```
SELECT * FROM rules /*We select all rules and we subtract from*/
MINUS
SELECT * FROM rules R /*them all rules included in at least*/
WHERE EXISTS (SELECT * FROM REGLES M /* another rule. A body of a*/
    WHERE M.id_r!=R.id_r AND EXISTS(SELECT item,income FROM rules S
/*rule X is included*/      WHERE R.income=S.income AND M.idr=S.idr
/*in another one Y if the*/            INTERSECT
/*intersection of X.body*/             SELECT item,income FROM rules P
/*and Y.body is not empty*/            WHERE R.idr=P.idr)
/* and if X is larger*/  AND (SELECT COUNT(*) FROM rules Q
/*than Y.*/                    WHERE Q.idr=M.idr)>
                              (SELECT COUNT(*) FROM rules U
                               WHERE U.idr=R.idr));
```

## 4.3. MINE RULE

MINE RULE is another DM Query Language. Its main query syntax is the following:
```
MINE RULE <RULETABLENAME> AS
SELECT DISTINCT [<CARDSPEC>] <ATTRBLIST> AS BODY,
                [<CARDSPEC>] <ATTRBLIST> AS HEAD [,SUPPORT][,CONFIDENCE]
[WHERE <WHERECLAUSE>]
FROM <FROMLIST> [WHERE <WHERECLAUSE>]
GROUP BY <ATTRBLIST> [HAVING <HAVINGCLAUSE>]
[CLUSTER BY <ATTRBLIST> [HAVING <HAVINGCLAUSE>]]
EXTRACTING RULES WITH SUPPORT:<REAL>, CONFIDENCE:<REAL>;
Avec
<CARDSPEC> = <NUMBER>..<NUMBER>|N.
```

This query enables association rule mining from data selected in the FROM clause and restricted by the WHERE clause. Rule body schema is defined in the SELECT clause by putting the attributes which corresponds to the rule body and their cardinality before the keyword BODY just as attributes, which correspond to rule's head and their cardinality before the keyword HEAD. Data are grouped thanks to the GROUP clause. Groups may be restricted by putting group conditions in the HAVING clause. The CLUSTER clause enables clustering data into groups. Then, rules are extracted only from within couples of clusters into the same group, one cluster for the body and the other for the head. For more details, refer to (Meo et al, 2002).

**Illustration**

We illustrate MINE RULE for the same Data Mining operations performed in the previous languages sections.

- **Pre-processing**

Data selection is possible since MINE RULE nests SQL. However, MINE RULE doesn't provide a discretization operator. Discretization is performed via the SQL function TRUNC(income,2) which rounds down income values. For example, incomes between 500 and 599 correspond to 500.

- **Association Rule Mining**

We formulate the appropriate MINE RULE query that mines association rules exceeding the support threshold 25% and the confidence threshold 50% and with bodies composed of items and heads of income.

```
MINE RULE TRANSACTION_RB AS
SELECT DISTINCT 1..N ITEM AS BODY, 1..1 TRUNC(INCOME,2) AS HEAD, SUPPORT, CONFIDENCE
FROM TRANSACTIONS T, CUSTOMER C
WHERE T.ID_CUSTOMER=C.ID_CUSTOMER AND PAYEMENT_MODE='credit_card'
GROUP BY ID_TRANSAC
EXTRACTING RULES WITH SUPPORT : 0.25, CONFIDENCE : 0.5 ;
```

Rule representation is lightly different form the DMQL one, but similar to the MSQL one in the way that rules are stored in a single table with four attributes; body and head that are vectors, support and confidence.

Table 3 : Table Transaction_RB

| Body | Consequent | Support | Confidence |
|---|---|---|---|
| {A} | TRUNC(income,2)=500 | 3 | 75% |
| {B} | TRUNC(income,2)=500 | 4 | 100% |
| {D} | TRUNC(income,2)=500 | 3 | 66% |
| {A,B} | TRUNC(income,2)=500 | 3 | 100% |

- **Post-processing**

Like DMQL, MINE RULE doesn't provide post-processing operations.

## 4.4. MineSQL

MineSQL is a SQL extension for data mining in relational databases. MineSQL queries aim mainly to mine association rules and also extract characteristic rules. The language provides a new data type called RULE, used to store and to manage rules. An attribute of type RULE has four components: set of body elements, set of head elements and support and confidence values.

There are several functions on the type RULE, such as the functions *support(r), confidence(r), body(r)* that return respectively the support, the confidence and the body of the rule *r*. There are also two operators *SATISFIED BY* and *VIOLATED BY*, which allow cross-over between data and patterns. MineSQL makes patterns post-processing easy via its functions and its two operators. The syntax of the main MineSQL query is the following:

```
MINE RULE_EXP [[AS] ALIAS]
[FOR {DATA_EXPR [USING TAX_NAME][AS ALIAS]|*}]
[TO {DATA_EXPR [USING TAX_NAME][AS ALIAS]|*}]
FROM TABLE [,TABLE]
[WHERE {DATA_CONDITION|RULE_CONDITIONG}
[{AND|OR} {DATA_CONDITION|RULE_CONDITION}]
[GROUP BY DATA_EXPR [,DATA_EXPR]
[HAVING CONDITION]]
[ORDER BY RULE_EXPR [{ASC|DESC}]]
```

In this query, *rule_exp* denotes the rule expression, i.e., the generated rule representation. *data_exp* denotes data expression, (attributes, constants, functions etc.). The structure of the rule is defined as a subset of attribute expressions in the FOR clause and the head is defined as a subset of attribute expressions in the TO clause. The clause FROM specifies data to be mined. The selection of extracted data and rules depends on conditions specified in the WHERE clause. The MINE statement inspects records, grouped by attributes indicated in the GROUP BY clause.

**Illustration**

We continue performing the same Data Mining operations via MineSQL queries.

- **Pre-processing**

Like DMQL and MINE RULE, MineSQL doesn't provide discretization query, but discretization may be done by specifying hierarchies (or taxonomy as called by some authors).

```
CREATE TAXONOMY INCOME_HIERARCHY(NODE 'AGE_DSCT', LEAF [MIN…499] REFERENCES 1, LEAF [500…599] REFERENCES 2, LEAF [600…699] REFERENCES 3, LEAF [700…799] REFERENCES 4, LEAF [800…MAX] REFERENCES 5);
```

- **Association Rule Mining**

We formulate MineSQL statement to extract association rules from transactions done by credit card, under support threshold 25% and confidence threshold 50%. Mined association rules are stored in a table we call *transaction_rb*. This table contains a column of type RULE and a column of type VARCHAR which is a description of the rule. The table creation statement is the following:

```
CREATE TABLE TRANSACTION_RB(RL RULE, DESCRIPTION VARCHAR(20));
```

The column RL is composed of a support value, a confidence value and of two sets of rule elements (Ai=v), where *Ai* is an item and *v* is a value of *Ai* domain; first set is the rule body, the second one is the head.

```
INSERT INTO TRANSACTION_RB(R)
MINE RULE, SUPPORT(RULE), CONFIDENCE(RULE)
FOR ITEM
TO INCOME USING INCOME_HIERARCHY AS INCOME_H
FROM (SELECT SET(ITEM), INCOME FROM TRANSACTIONS T, CUSTOMER C
    WHERE T.ID_CUSTOMER=C.ID_CUSTOMER AND PAYMENT_MODE='credit_card'
    GROUP BY T.ID_TRANSAC, C.INCOME)
WHERE SUPPORT(RULE)>0.25 AND CONFIDENCE(RULE)>0.5
```

Figure 7 shows *transaction_br* table. Note that it contains only two columns.

- **Post-processing**

MineSQL allows some post-processing operations, in particular for cross-over patterns and data. We achieve the same operation done previously, which is selecting data that violate association rules implicating the item A:

```
SELECT ID_TRANSAC FROM TRANSACTION T1
WHERE (SELECT * FROM TRANSACTION_RB TBR WHERE 'ITEM='A'' IN BODY(TBR.RGL))VIOLATED BY(SELECT SET(ITEM),
INCOME FROM TRANSACTION T2, CUSTOMER C
WHERE T2.ID_CUSTOMER=C.CUSTOMER AND T1.ID_TRANSAC=T2.ID_TRANSAC
GROUP BY T2.ID_TRANSAC, C.INCOME);
```

Figure 4: Table transaction_rb

| Rule | s | c | Description |
|---|---|---|---|
| item='A'→income_h=2 | 0.37 | 0.75 | |
| item='B'→income_h=2 | 0.5 | 1.0 | |
| item='D'→income_h=2 | 0.37 | 0.66 | |
| item='A' & item='B'→income_h=2 | 0.37 | 1.0 | |

The transaction identifiers selected by the query above are: t2, t4, t5, t9 and t10. The language also allows selecting rules with maximal bodies. We use MineSQL functions to extract it:

```
SELECT * FROM TRANSACTION_RB TRB1
WHERE NOT EXISTS(SELECT * FROM TRANSACTION_RB TRB2
WHERE BODY(TRB1.RGL) IN BODY(TRB2.RGL)
AND HEAD(TRB1.RGL)=HEAD(TRB2.RGL));
```

The selected association rules are:
```
Item='D'→Income=[500,599] and Item='A'&item='B'→Income=[500,599].
```

## 4.5. DMX

The OLE DB for Data Mining API defines common data mining concepts such as mining models, model training, model content, model prediction and so on. It also defines a query language for data mining. The syntax of this query language is similar to SQL (Tang and MacLennan, 2005). A data mining model or mining model can be thought of as a relational table. It contains key columns, input columns and predictable columns. Each model is associated with a data mining algorithm, on which the model is trained. After training, the data mining model stores the patterns discovered by the data mining algorithm about the dataset. While a relational table is a container of records, a data mining model is a container of patterns.

So a model is created; it deals with creating an empty data mining model, similar to the way we create a new table. Then, it's trained; the model training is used to invoke the data mining algorithm to uncover knowledge about the training dataset. After training, the patterns are stored in the mining model. Finally, we can apply the trained model to the new dataset and predict potential values of predictable columns for each new case: this is called model prediction (Tang and MacLennan, 2005). This is a simplified DMX syntax query to create a model:
```
CREATE MINING MODEL <model>
(<comma-separated list of column definitions>)
USING <algorithm> [()]
```
And this is a simplified DMX syntax query to train a model:
```
INSERT INTO <model> (<mapped model columns>) <source data query>
```
**Illustration**
- **Pre-processing**

DMX is different from the other languages because data selection is done after model creation. In Model Creation step, the data miner has to select only the attributes concerned with the model.

Discretization is the second pre-processing operation we do in this phase. DMX provides four statistical and data mining methods to do that: clusters, equal areas, thresholds and automatic (see (Tang and MacLennan, 2005)).

- **Association Rule Mining**

In fact, the two steps (pre-processing and ARM) are not really separated for DMX. For example discretization is done on the fly, and data selection is done after model definition.
```
CREATE MINING MODEL TRANSACTION_RB (ID_TRANSAC LONG KEY
INCOME LONG DISCRETIZED PREDICT_ONLY,
ITEMS TABLE(ITEM TEXT KEY))
USING MICROSOFT_ASSOCIATION_RULES (MINIMUM_SUPPORT=2, MIN_PROBABILITY=0.5)
```
In an association model, if a column is used for input, its values can be used only on the left side of association rules. If a column is used to make predictions, the column's states can be used on the left and right sides of the association rules. If a column is PREDICT_ONLY, it appears on the right side of rules. In the model creation query, our model has two attributes. The first one is INCOME, which must be discretized; it's a PREDICT_ONLY variable since we want it in the rule head. The second model attribute is a table we call ITEMS which represents the set of items bought by customers. This attribute is used for input (only in rule body).

While defining algorithm parameters and syntactic constraints is called model creation, association rules generation following our model definition is the model training:
```
INSERT INTO [TRANSACTION_RB]
(ID_TRANSAC,INCOME,ITEMS)
SHAPE
{SELECT [ID_TRANSAC],[INCOME]
FROM TRANSACTIONS A,CUSTOMER
WHERE A.ID_CUSTOMER=CUSTOMER.ID_CUSTOMER
AND A.PAYMENT_MODE="credit_card"
APPEND(
{SELECT [ITEM] FROM TRANSACTIONS B
RELATE A.ID_TRANSAC TO B.ID_TRANSAC
ORDER BY [ID_TRANSAC]})
AS [ITEMS]}
```
Here, the keyword SHAPE is used to specify target data; in deed we must select each transaction with its basket. The keyword APPEND is used to achieve this aim and to complete the data selection. It is used to add the set of items bought in for each transaction. The keyword RELATE is used to put the convenient basket to each transaction.

- **Post-processing**

The OLE DB for DM model has a representation tree with three levels. The top level has a single node that represents the model. The second level contains nodes representing qualified itemsets with their associated supports. The rowsets of the itemset nodes contain detailed information about the itemsets, with each row representing an individual item. The third level contains nodes that represent qualified rules. The parent of the rule node is the itemset that represents the left side item of a rule. The right side of a rule always has a single item, which is stored in the rowsets.

To select all retrieved association rules, we query the node type 8; the node type 7 is about itemsets:
```
SELECT NODE_DESCRIPTION FROM TRANSACTION_RB.CONTENT
WHERE NODE_TYPE=8;
```
We note that association rules are stored in PMML format which is XML-based. OLE DB for DM doesn't provide real operators or functions to post-process association rules like MSQL and MineSQL.

## 4.6. ODMQL

ODMQL allows the user to formulate data mining queries in an OQL-like syntax (Elfeky et al, 2000). The Data Mining language supports specification of four rule types: characteristic rules, discriminant rules, classification rules and association rules. The set of relevant data is specified by the WITH RELEVANCE TO clause. ODMQL also supports the specification of some kinds of threshold according to the mined rule type and allows the user to specify concept hierarchies for the attributes of the schema. Concept hierarchies could be created, modified and dropped. The general syntax of ODMQL queries (mining rules and manipulating concept hierarchies) is the following:

```
MINE FOR <rule_specification>
WITH RELEVANCE TO <projection_attributes>
FROM <variable_declaration> {,<variable_declaration>}
[WHERE <condition>]
[WITH THRESHOLD[S](<threshold>,{<threshold>})]
```

Here, rule_specification could be ASSOCIATION RULES, CHARACTERISTIC RULES, DISCRIMINANT RULES or CLASSIFICATION RULES. Threshold is specified in this form:

```
<threshold>::=<threshold_type>=<numerical_value>
```

The syntax used to define concept hierarchies is:

```
DEFINE HIERARCHY FOR <attribute>
ANY -> <concept_set> {,<concept_definition>}
<concept_definition>::=<concept>-><concept_set>
<concept_set>::={<concept>{,<concept>}}|[numerical_value..numerical_value]
<concept>::=ANY|<string_lateral>
```

**Illustration**

- **Pre-processing**

ODMQL nests OQL, so it allows data selection via the FROM and WITH clauses. Unfortunately, it doesn't provide a discretization operator, but we can use concept hierarchies to define all possible values for each income interval, i.e., income discretization.

```
DEFINE HIERARCHY FOR INCOME:
ANY-> {I4,I5,I6,I7,I8},
I4->[0..499],I5->[500..599],I6->[600..699],I7->[700..799],I8->[800..Int_Limit];
```

- **Association Rule Mining**

In deed, we didn't find any track of association rule mining syntax or an example for extracting association rules that are multidimensional (items in the rule correspond to different attributes) for ODMQL language. That's why we cannot illustrate the mining operator of ODMQL in our example.

- **Post-processing**

ODMQL doesn't provide any post-processing operator. This primitive remains the most ignored by Data Mining Languages despite its importance and its definition as a primitive.

## 5. DISCUSSION

In this paper, we studied existent DM Languages extending the work of (Botta et al, 2004). The objective is to get leverage on achievements and limits of the state of the art. We note that generally, SQL is nested in almost all languages enhancing the data selection capabilities. However, the pre-processing and transformation KDD steps are not supported by the whole of the language. MSQL and DMX are the only languages that provide a discretization operator. In these preliminaries KDD steps, there is also a need for sampling techniques, cleaning techniques, errors and missing values handling techniques etc.

The step that follows data pre-processing is the DM. The DM step is provided by all languages. All of them focus on mining patterns from a dataset. DMX, DMQL and ODMQL offer to the Data Miner several kinds of patterns, which constitute an important feature for an exhaustive Data Mining language. In addition, background knowledge specification is provided by almost all the languages except MSQL.

Regarding post-processing, there is a great lack for this KDD step. Only MSQL and MineSQL provide operators and functions allowing the data miner to handle and to manipulate patterns. The same languages also provide operators to cross-over data and patterns. Finally, the whole of the languages satisfy the closure property since all patterns are stored in relational database (except ODMQL). The closure property allows complex manipulation and selection of data and/or patterns enhancing the capabilities of the language.

A crucial step for a standard DM query language is to define clearly its primitives. A better understanding of Data Miners requirements and DM challenges (Fayyad et al, 2003) is the key feature of a good primitives' definition. Then, a theory of inductive query answering is required to formulate all KDD operations and properties in an algebraic language. Several researchers tried to do so ((Boulicaut et al, 1999), (De Raedt, 2003), (Dan Lee and De Raedt, 2003) and (De Raedt et al, 2004)) and aimed to answer inductive queries. Once primitives are better defined and an inductive query evaluation theory is clearly formulated, computer science techniques of computation and optimization will evolve and advances will be achieved, especially in constraint-based mining and optimization of query evaluation (Jeudy and Boulicaut, 2002), (Pensa et al, 2006). A practical and efficient DM query language should produce an exponential emergence of DM applications and inductive databases. There are surely many inductive databases ((Han et al,1996b), (Kaufman and Michalski, 2002) and (Helma et al, 2002)) and DM applications in the nature, but all of them present isolated functionalities due to the lack of a standard framework and theory of DM.